\documentclass{article}

     \PassOptionsToPackage{numbers, compress}{natbib}

    \usepackage[preprint]{neurips_2019}

\usepackage[utf8]{inputenc} 
\usepackage[T1]{fontenc}    
\usepackage{hyperref}       
\usepackage{url}            
\usepackage{booktabs}       
\usepackage{amsfonts}       
\usepackage{nicefrac}       
\usepackage{microtype}      
\usepackage{graphicx}
\usepackage{amsmath}
\usepackage{booktabs}
\usepackage{adjustbox}
\usepackage{paralist}
\usepackage{amssymb}
\usepackage{subcaption}

\newcommand\numberthis{\addtocounter{equation}{1}\tag{\theequation}}
\newcommand*{\QEDB}{\hfill\ensuremath{\square}}

\title{A Deep neuro-fuzzy network for image classification}

\author{
Omolbanin Yazdanbakhsh 
  \\
  Xsensor Technology Corp\\
  Calgary, AB, Canada \\
  \texttt{yazdanba@ualberta.ca} \\
 \And
 Scott Dick \\
 Department of Electrical and Computer Engineering \\
 University of Alberta \\
 Edmonton, AB, Canada \\
 \texttt{scott.dick@ualberta.ca} \\
 }

\begin{document}

\maketitle

\begin{abstract}
 The combination of neural network and fuzzy systems into neuro-fuzzy systems integrates fuzzy reasoning rules into the connectionist networks. However, the existing neuro-fuzzy systems are developed under shallow structures having lower generalization capacity.  We propose the first end-to-end deep neuro-fuzzy network and investigate its application for image classification. Two new operations are developed based on definitions of Takagi-Sugeno-Kang (TSK) fuzzy model namely fuzzy inference operation and fuzzy pooling operations; stacks of these operations comprise the layers in this network. We evaluate the network on MNIST, CIFAR-10 and CIFAR-100 datasets, finding that the network has a reasonable accuracy in these benchmarks.
 \end{abstract}

\section{Introduction}

Performance of many real-world problems have been significantly improved by replacing shallow structures with deeper networks. Various problems such as image classification \citep{krizhevsky2012imagenet}, object detection \citep{ren2015faster}, semantic segmentation \citep{noh2015learning}, and sequence modeling \citep{gehring2017convolutional} had their breakthroughs in recent years by employing deep neural networks. Basically, deep neural networks are stacks of multiple hidden layers or classifiers extracting complex features from inputs; the networks integrate low-, mid-, and high-level features of different layers to improve generalization and performance \citep{bengio2009learning}. 

Neuro-fuzzy systems are referred to models working based on a combination of neural network and fuzzy systems. In general, in these models, designing of a fuzzy system follows learning procedures of neural networks \citep{nauck2013neuro}. Neuro-fuzzy networks are usually applied to function approximation problems such as classification tasks and control systems \citep{pedrycz1998neurofuzzy}. Various structures of neuro-fuzzy have been proposed \citep{nauck2013neuro}; the most popular neuro-fuzzy model is a 5-layer feed-forward network called Adaptive neuro fuzzy inference system (ANFIS) where parameters of a TSK fuzzy model were calculated based on a neural network framework \citep{jang1993anfis}. The proposed structures for neuro-fuzzy models are shallow as they only model one rule-set of a fuzzy system. Likewise other learning structures, introducing deeper structures in neuro-fuzzy models may improve their performances.

Deep structures have been introduced in neuro-fuzzy frameworks in recent years by chaining stacks of TSK systems \citep{rajurkar2017developing,zhang2018deep,zhou2017stacked,zhou2017deep}. Outputs of each level in these models remain in the same space as their original input space throughout the networks using random shift \citep{zhou2017deep} and feature augmentation \citep{zhou2017stacked,zhang2018deep}; the models have been applied for classification of feature-based datasets. For image classification, several works have considered combination of deep neural networks and fuzzy systems \citep{gu2018semi,yeganejou2018classification,riaz2019semi}; in these models, a fuzzy clustering or a fuzzy rule-based system is applied on the features extracted using a well-known deep neural network. However, none of the networks have explored using a deep neuro-fuzzy network as an end-to-end network for image classification.

In this paper, we propose an end-to-end deep neuro-fuzzy network for image classification. Layers in this network are stacks of two new operations developed based on TSK system concepts; we call the operations fuzzy inference and fuzzy pooling operation. To the best of our knowledge, the proposed network is the first end-to-end deep neuro-fuzzy structure for image classification. Experiments on MNIST \citep{lecun1998gradient}, CIFAR-10 and CIFAR-100 \citep{krizhevsky2009learning} demonstrate that the model has comparable performance for image classification tasks.

The remainder of the paper is organized as follow. In section 2, we provide background on neuro-fuzzy systems and review latest works on deep neuro-fuzzy structures. Our proposed network is presented in Section 3, and our experimental result are in Section 4. We close with a summary and discussion of future works in Section 5.

\section{Background Review}

\subsection{Fuzzy Logic and Neuro-Fuzzy Structures}

A fuzzy set is defined as a membership function mapping elements of a universe of discourse (X) to the unit interval [0,1] \citep{zadeh1996fuzzy}.

\begin{equation} \label{eq:mapping-unit}
A:X \to [0,1]
\end{equation}

For data analysis applications, membership grade can be viewed as a degree of similarity, preference and uncertainty. From similarity perspective, a membership grade can represent degree of compatibility of an element in the universe of discourse $(x \in X)$ with representative elements of A \citep{pedrycz2007fuzzy}. 

Fuzzy if-then rules are expression with the form of IF A THEN B where A is a fuzzy set and B is either a fuzzy set or a function of inputs. The fuzzy rules aim to add human-level decision making procedure into a system to capture uncertainty and imprecision of environment. A fuzzy rule-based system is built of multiple fuzzy rules where each of if-then fuzzy rules works as a local descriptor of environment \citep{angelov2010simple,jang1993anfis}.

Neuro-fuzzy systems are architectures to model fuzzy rule-based systems. ANFIS is one of the first and most popular neuro-fuzzy architectures proposed in 1993 \citep{jang1993anfis}; it is a layered feed-forward network based on TSK inferential system. A hybrid of gradient descent and least-square estimation is used to learn the network parameters. A TSK rule set is defined as \citep{jang1993anfis}:

\begin{equation} \label{eq:tsk}
\textrm{If } x_1 \textrm{ is } A_{1k} \textrm{ and } x_2 \textrm{ is } A_{2k} \textrm{ And ... } x_d \textrm{ is } A_{dk} \textrm{ then } y_k=f_k(x)
\end{equation}

where $k=1,2,...,K$ and $K$ is the number of fuzzy rules, $x_i$ is the $i$th input variable, $A_{ik}$ is a fuzzy set for $i$th input of $k$th rule, $And$ is a fuzzy conjunction operator, and $f_k (.)$ is output of  $k$th rule. The output of the system is calculated as

\begin{equation} \label{eq:output}
y = \frac{\sum_{k=1}^{K} p_k(x).f_k(x)}{\sum_{k=1}^{K} p_k(x)}=   \sum_{k=1}^{K} \bar{p}_k(x).f_k(x)
\end{equation}

where $p_k(x) = \prod_{i=1}^d p_{A_{ik}}(x_i)$ and $p_{A_{ik}}(x_i)$ is a membership grade measuring degree of similarity between $x_i$ and $A_{ik}$. Each layer in ANFIS network is described as follows \cite{jang1993anfis}:

\begin{itemize}
  \item \textit{Layer 1}: Membership grade of an input is calculated as: $p_{A_{ik}}$ 	
  \item \textit{Layer 2}: Firing strength of kth rule is obtained as:  $p_k(x) = \prod_{i=1}^d p_{A_{ik}}(x_i)$
  \item \textit{Layer 3}: Firing strength of each rule is normalized as: $\bar{p}_k(x) = \frac{p_k(x)}{\sum_{j=1}^K p_j(x)}$ 	
  \item \textit{Layer 4}: The output of each rule is calculated as: $\bar{p}_k(x).f_k(x)$
  \item \textit{Layer 5}: Final output of the network is obtained as: $\sum_{k=1}^K \bar{p}_k(x).f_k(x)$
\end{itemize}

\subsection{Deep neuro-fuzzy Structures}

There is a small amount of works integrating fuzzy logic and deep learning algorithms. Aviles et.al \citep{aviles2016deep} combined ANFIS and Long-Short Term Memory (LSTM) structure for estimation of the interaction forces in robotic assisted minimally invasive scenarios. \citep{chen2015fuzzy} proposed a fuzzy restricted Boltzmann machine (FRBM) where parameters in the model are fuzzy numbers. \citep{zheng2017airline} extended FRBM with Pythagorean fuzzy numbers \citep{yager2013pythagorean} and applied the model for airline passenger profiling, and \citep{shukla2017interval} extended FRBM with interval Type-2 fuzzy numbers \citep{mendel2002type}. \citep{zheng2017pythagorean} used Pythagorean fuzzy values to express distribution of parameters in a deep denoising auto-encoder and applied it for early warning of industrial accident. \citep{rajurkar2017developing,zhang2018deep,zhou2017stacked,zhou2017deep} developed deep fuzzy structures using stack of TSK fuzzy systems. \citep{rajurkar2017developing} considered each node in a layer as an independent TSK model. \citep{zhou2017stacked} fed a random shift of outputs in the previous fuzzy system along with original input as input to the next fuzzy system. \citep{zhang2018deep,zhou2017deep} augmented output of the previous fuzzy system to the original input and fed it as input to the next fuzzy system. \citep{pratama2018incremental} proposed an evolving deep neuro-fuzzy structure for studying dynamic data streams.

Few studies have considered integration of fuzzy systems and deep learning for image classification \citep{gu2018semi,yeganejou2018classification,riaz2019semi}. In \citep{gu2018semi}, first, features were extracted from images using a VGGNet \citep{simonyan2014very}, and then a set of fuzzy rule-based layers \citep{angelov2010simple} were applied to classify the images. \citep{yeganejou2018classification} classified images by applying a fuzzy c-means clustering on the features extracted from a CNN. \citep{riaz2019semi} uses a combination of features extracted from CNN and fuzzy rough c-means clustering for semi-supervised image classification.

\section{Proposed Structure}

To develop an end-to-end deep neuro-fuzzy network for image classification, we introduce two operations namely fuzzy inference operation and fuzzy pooling operation. Stacks of the operations comprise the network. In this section, we review design of these operations and our deep neuro-fuzzy network.

\subsection{Fuzzy Inference Operation}

A Fuzzy inference operation models a TSK rule-based system for image analysis. To develope this operation, we work with single-input-multi-output (SIMO) rule-based systems as

\begin{align*}
\textrm{If } x \textrm{ is } A_k & \textrm{ then } \begin{pmatrix}f_{1k}(x) \\  f_{2k}(x) \\ \vdots \\  f_{nk}(x)\end{pmatrix},    k=1,...,K \numberthis  \label{eq:fuzzy-conv}
\end{align*}

where $x$ is input variable from universe of discourse ($X$), $K$ in number of rules, $n$ is number of outputs, $A_k$ is a fuzzy set for the $k$th rule defined over $X$, and $f_{ik}$ is the $i$th output of $k$th rule; the outputs are nonlinear functions of inputs. To employ this set of rules for image analysis, we follow fuzzy rule-based method proposed in \citep{angelov2010simple} where premise of the rules can present statements such as ``$x$ is like $A_k$''. 

For image analysis, we consider the universe of discourse as a set of subregions of a given image. Also, we consider a fuzzy set to be a  specific pattern. Finally, we consider the membership grade to be the similarity between the subregions and the given pattern. Each fuzzy rule in the proposed rule set captures a different pattern in the image. Semantically, we can rewrite Equation \ref{eq:fuzzy-conv} 
as:

\begin{align*}
\textrm{If subregions in } X \textrm{ is similar to } & \textrm{pattern specified by } A_k \textrm{ then } \begin{pmatrix}f_{1k}(x) \\  f_{2k}(x) \\ \vdots \\  f_{nk}(x)\end{pmatrix},    k=1,...,K   \numberthis \label{eq:fuzzy-conv2}
\end{align*}

Thus, the proposed model follows a SIMO rule-based system where the single input is the given image, $k$th rule captures a pattern in the subregions of the image, each output ($f_{ik}(x)$) presents a nonlinear function over the subregions for the $k$th pattern, and the defuzzification outputs are impacted by combination of patterns specified by fuzzy sets in each rule on the image. 

A fuzzy inference operation includes the following three steps.

\begin{enumerate}
  \item 	\textit{Calculating membership matrix}: For each rule, a matrix of membership grades ($P_k$) is calculated where each item in the matrix shows similarity between a subregion in the image, and a fuzzy set ($A_k$). We employ dot product operation for similarity measurement, and the membership values ($p_{ik}$) are restricted to be in $[0,1]$.
  \begin{equation} \label{eq:mf-matrix}
{P_k}=[p_{ik}]=[a_i \cdot A_k]
\end{equation}
where $a_{i}$ is $i$th subregion of the image, $\cdot$ shows dot product, and $p_{ik}$ is $i$th item in $P_k$.
  \item \textit{Calculating firing strength}: Firing strength of each rule is calculated by normalizing the membership grade matrix as: 
\begin{equation} \label{eq:rule-strength}
\overline{P_k}=\frac{P_k}{\sum_{j=1}^K P_j}
\end{equation}
where $K$ is number of rules.
  \item	\textit{Calculating final outputs}: Final outputs of the rule-based system are obtained as
\begin{equation} \label{eq:rule-output}
y_i=\sum_{k=1}^K y_{ik}=\sum_{k=1}^K \overline{P_k}f_{ik}(x)
\end{equation}
where $y_i$ is $i$th output and $y_{ik}$ is $i$th output in $k$th rule.
\end{enumerate}
Fuzzy sets (patterns) and parameters of the output function ($f_{ik} (x)$) are learned using gradient descent.

\subsubsection{Example 1 - Fuzzy Inference Operation for a $4 \times 4$ image}

The steps for a fuzzy inference operation are described in details in this example. The input is a $4 \times 4$ one-channel image ($a$) and the fuzzy set is a $2 \times 2$ pattern as $w = \begin{pmatrix} w_{1k} & w_{2k}  \\  w_{3k} & w_{4k}  \end{pmatrix}$ where $k$ shows the corresponding fuzzy rule.

\subsubsubsection{Calculating membership matrix}

\begin{figure*}[ht]
\vskip 0.2in
\begin{center}
\centerline{\includegraphics[scale=0.5]{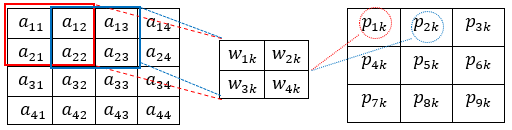}}
\caption{Membership matrix calculation}
\label{fig:mf_calc}
\end{center}
\vskip -0.2in
\end{figure*}

First, the universe of discourse is created by dividing the image to subregions whose size is same as fuzzy sets; we consider $2 \times 2$ subregions with stride of 1. For each rule, a membership matrix is obtained by calculating similarity between each subregion and the pattern specified by its fuzzy set using dot product as
\begin{equation} \label{eq:dot_proc}
p_{ik}=a_1w_{1k}+a_2w_{2k}+a_3w_{3k}+a_4w_{4k}
\end{equation}
where $p_{ik}$ is the $i$th item in the membership matrix of $k$th rule and shows similarity between a subregion,  $\begin{pmatrix} a_1 & a_2  \\  a_3 & a_4  \end{pmatrix}$ , and the pattern specified with the fuzzy set $A_k$,  $\begin{pmatrix} w_{1k} & w_{2k}  \\  w_{3k} & w_{4k}  \end{pmatrix}$. Figure \ref{fig:mf_calc} shows the membership grade assignment for the image.

Calculating the membership matrix for each rule is similar to calculating convolution between an image and a filter; in this example, the operation resembles a convolution between a $4 \times 4$ image with a filter of size $2 \times 2$ with stride of 1. The main difference is that the values in the membership matrix are in range [0,1]. 

To implement membership matrix in the proposed fuzzy inference operation, we use a convolution operation where number of filters is same as number of rules, and size of filters is equal to the size of patterns specified by the fuzzy sets. As the membership grades are between [0,1], the output of the operation is clipped to be between [0,1]. 

\subsubsubsection{Calculating firing strength}

To calculate firing strength of each rule, each item in the membership matrix of $k$th rule is normalized by its corresponding item in the other membership matrices, i.e. $\overline{p_{ik}}=\frac{p_{ik}}{\sum_{k=1}^K p_{ik}}$ where $\overline{p_{ik}}$ is $i$th item in firing strength, $\overline{P_k}$, of $k$th rule.

\subsubsubsection{Calculating final output}

\begin{figure*}[ht]
\vskip 0.2in
\begin{center}
\centerline{\includegraphics[scale=0.5]{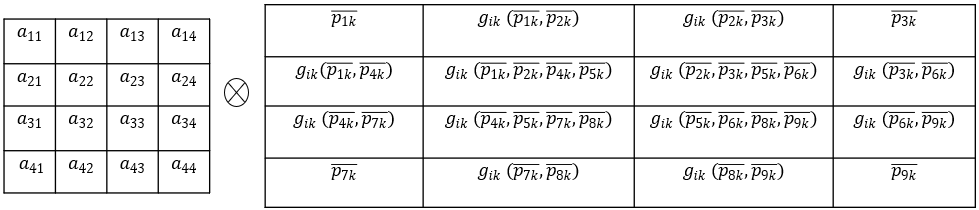}}
\caption{Element-wise multiplication between image and $g_{ik}(\overline{P_k})$}
\label{fig:gk_image}
\end{center}
\vskip -0.2in
\end{figure*}

The following shows calculations for the $i$th output of the $n$ outputs ($y_j ,j=1,…,n$). Output of each rule is calculated as  $y_{ik}=\overline{P_k } f_{ik}(x)$; each item in $\overline{P_k }$ is multiplied only to its corresponding subregion. Figure \ref{fig:firing_calc} shows the relation between subregions and the firing strength. Thus, the output is obtained as:

\begin{figure*}[ht]
\vskip 0.2in
\begin{center}
\centerline{\includegraphics[scale=0.5]{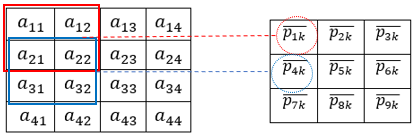}}
\caption{Relation between subregions and the firing strength}
\label{fig:firing_calc}
\end{center}
\vskip -0.2in
\end{figure*}

\begin{equation} \label{eq:final_out}
y_{ik}=\begin{pmatrix} a_{11}.\overline{p_{1k}} & a_{12}.g_{ik}(\overline{p_{1k}},\overline{p_{2k}}) &a_{13}.g_{ik}(\overline{p_{2k}},\overline{p_{3k}}) &a_{14}.\overline{p_{3k}} \\  a_{21}.g_{ik}(\overline{p_{1k}},\overline{p_{4k}}) & a_{22}.g_{ik}(\overline{p_{1k}},\overline{p_{2k}},\overline{p_{4k}},\overline{p_{5k}}) &a_{23}.g_{ik}(\overline{p_{2k}},\overline{p_{3k}},\overline{p_{5k}},\overline{p_{6k}})&a_{24}.g_{ik}(\overline{p_{3k}},\overline{p_{6k}})\\   a_{31}.g_{ik}(\overline{p_{4k}},\overline{p_{7k}}) & a_{32}.g_{ik}(\overline{p_{4k}},\overline{p_{5k}},\overline{p_{7k}},\overline{p_{8k}}) &a_{33}.g_{ik}(\overline{p_{5k}},\overline{p_{6k}},\overline{p_{8k}},\overline{p_{9k}})&a_{34}.g_{ik}(\overline{p_{6k}},\overline{p_{9k}}) \\  a_{41}.\overline{p_{7k}} & a_{42}.g_{ik}(\overline{p_{7k}},\overline{p_{8k}})  &a_{43}.g_{ik}(\overline{p_{8k}},\overline{p_{9k}}) &a_{44}.\overline{p_{9k}} \end{pmatrix}
\end{equation}
where $g_{ik}()$ is a function defined to deal with overlapping subregions. The output can be considered as an element-wise multiplication between  $g_{ik}(\overline{P_k})$ and $f_{ik}(x)$; note that in this example to simplify visualization, we consider $f_{ik}(x)$ as an identity function on the subregions. Figure \ref{fig:gk_image} shows this process.

\begin{figure*}[ht]
\vskip 0.2in
\begin{center}
\centerline{\includegraphics[scale=0.5]{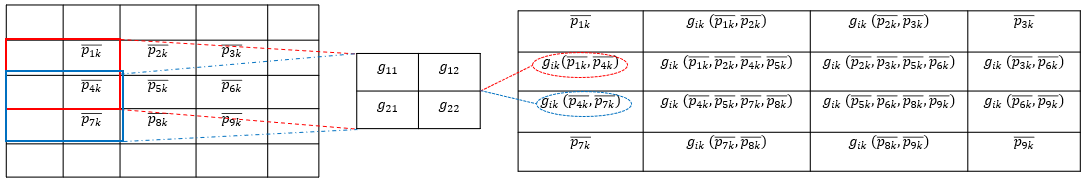}}
\caption{ $g_{ik}(\overline{P_k})$ calculation using convolution operation}
\label{fig:gk_conv}
\end{center}
\vskip -0.2in
\end{figure*}

$g_{ik}(\overline{P_k})$ resembles a convolution operation between a $2 \times 2$ filter with stride 1 and a padded version of $\overline{P_k}$ which is padded in height and width with a length as the difference between the image size and the membership grade matrix. In this example, $\overline{P_k}$ is padded with $image\_size-membership\_matrix\_size=4-3=1$. Figure \ref{fig:gk_conv} demonstrates $g_{ik}(\overline{P_k})$ calculation

Therefore, output of each fuzzy rule is calculated using the following three steps: \begin{inparaenum} \item Pad $\overline{P_k}$ \item Convolve the padded $\overline{P_k}$ with a filter with the same size as the fuzzy sets \item Do an element-wise multiplication between the image and the result of step 2. \end{inparaenum}To obtain the final output, $y_i$, we sum $y_{ik}$ over the rules. \QEDB

\begin{figure*}[ht]
\vskip 0.2in
\begin{center}
\centerline{\includegraphics[width=0.9\textwidth, scale=0.25]{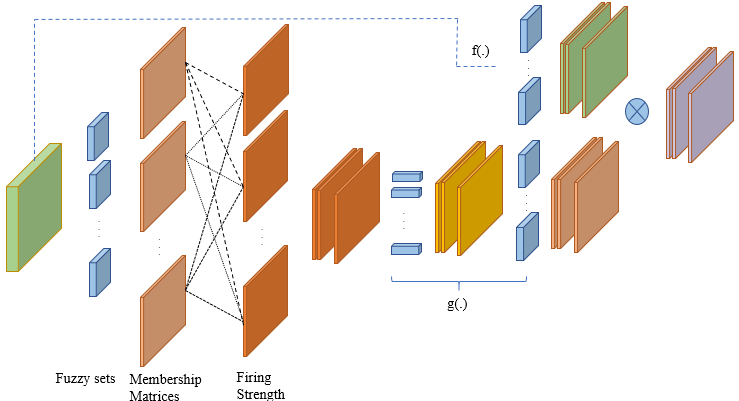}}
\caption{fuzzy inference operation for TSK rule-based system with multiple outputs}
\label{fig:convTsk}
\end{center}
\vskip -0.2in
\end{figure*}

We can implement the proposed fuzzy inference operation efficiently in machine learning frameworks such as Tensorflow \citep{abadi2016tensorflow}.  To do so, Equation \ref{eq:rule-output} is modified as $y_i=g_i(\overline{P_1},\overline{P_2},...,\overline{P_k}).f_i (x)$. Table \ref{tab:fuzzyConvOp} shows steps to calculate a fuzzy inference operation for multi-output TSK rule-based system and Figure \ref{fig:convTsk} demonstrates diagram of the fuzzy inference operation.

\begin{table*}[tb]
\caption{Steps for a fuzzy inference operation}
\label{tab:fuzzyConvOp}
\vskip 0.15in
\begin{center}
\begin{small}
\begin{tabular}{lcccccr}
\toprule
1. Obtain membership matrix of each rule by convolving a filter with the image. \\
2. Clip the values in the membership grade to be in the range of [0,1]. \\
3. Obtain firing strength of each rule by normalizing membership grade using Equation \ref{eq:rule-strength}. \\
4. Concatenate membership matrices as depth of a matrix as $F_1=\begin{pmatrix} \overline{P_1} & \overline{P_2} & ... &\overline{P_K} \end{pmatrix}$. \\
5. Pad $F_1$ based on size difference between the image and the membership matrix, and call it $F_2$. \\
6. To implement $g(.)$, first, a 1*1 filter is convolved with $F_2$ to get the relation of corresponding items in \\ \-\ \-\ $\overline{P_1}, \overline{P_2}, \cdot, \overline{P_K}$. Then, the obtained featured maps are convolved with filters of the same size as the patterns. \\ \-\ \-\ To apply this step for all the outputs, $y_i$,  simultaneously, number of filters in this step is considered as the \\ \-\ \-\  number of outputs, $n$. We call the output of this step $F_3$. \\
7. Convolve image with a filter with the same size as fuzzy sets to implement $f(x)$ and call it $F_4$; this \\ \-\ \-\ operation also helps us to work with multi-channel inputs. Number of filters is equal to the number of outputs. \\
8. Do an element-wise multiplication between $F_4$ and $F_3$. \\
\bottomrule
\end{tabular}
\end{small}
\end{center}
\vskip -0.1in
\end{table*}

\subsection{Fuzzy Pooling Operation}

In the fuzzy inference operation, input and output have the same size because number of subregions defined for an image is equal to number of pixels in the image. If we consider fewer number of subregions, we can reduce the size of outputs. For example, for a $4 \times 4$ image, if we have 4 subregions, we can reduce the size of outputs to $2 \times 2$; in other words, the outputs have size of  $\sqrt{\#\textrm{ of subregions}}$. Implementation of fuzzy pooling operation is similar to fuzzy inference. The main difference is that there is no need for the step 5 and part of the step 6 in Table \ref{tab:fuzzyConvOp}. Table \ref{tab:poolingOp} shows the steps for fuzzy pooling operation.

\begin{table*}[tb]
\caption{Steps for fuzzy Pooling operation}
\label{tab:poolingOp}
\vskip 0.15in
\begin{center}
\begin{small}
\begin{tabular}{lcccccr}
\toprule
1. Calculate membership matrices as the steps 1 - 4 in Table \ref{tab:fuzzyConvOp}; we call the output $F_1$. \\
2. $1\times 1$ filers are convolved with $F_1$ and called $F_2$. \\
3. Convolve image with filters with the same size as fuzzy sets to implement $f(x)$ and call it $F_3$. \\
4. Do an element-wise multiplication between $F_2$ and $F_3$. \\
\bottomrule
\end{tabular}
\end{small}
\end{center}
\vskip -0.1in
\end{table*}

\subsection{Deep neuro-fuzzy network}

To build a deep neuro-fuzzy, we chain stacks of fuzzy inference operation and fuzzy pooling operation. Input to an operation is outputs of a previous operation. For example ,$y_i,   i=1,…,n$ are concatenated in a matrix and passed as input to the next operation.

The proposed network is able to extract local and global features for image analysis. For image classification, the extracted features are passed to fully connected layers for classification task. The network is trained end-to-end using gradient descent. Parameters of filters in fuzzy inference and fuzzy pooling operations along with fully connected layers parameters are learned. A detailed description of learnable parameters of fuzzy inference operation based on the steps of Table \ref{tab:fuzzyConvOp} is as follow:

For step 1, filters related to fuzzy sets are learnable; from step 2-5 there is no parameter. In step 6, we have parameters for $1 \times 1$ filters, also we consider Leaky-Relu \citep{xu2015empirical} as nonlinearity; for second part of layer 6, we use average pooling instead of convolution operations. In step 7, we have learnable parameters and Leaky-Relu is deployed as nonlinearity, and there is no parameter in step 8.

\section{Experiments}

In this section, we apply deep neuro-fuzzy networks for classification of MNIST \citep{lecun1998gradient}, CIFAR10 and CIFAR-100 \cite{krizhevsky2009learning} datasets. 

\begin{figure*}[ht]
\vskip 0.2in
\begin{center}
\centerline{\includegraphics[width=\textwidth]{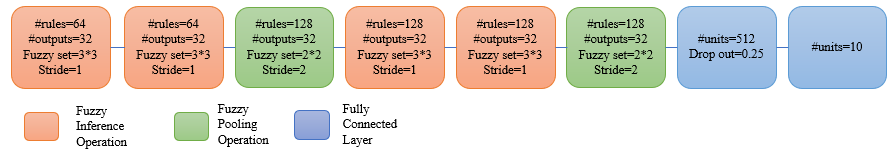}}
\caption{Network designed for MNIST dataset}
\label{fig:mnistNetwork}
\end{center}
\vskip -0.2in
\end{figure*}

MNIST \citep{lecun1998gradient} consists handwritten digits of 0 to 9. The dataset consists of 60k training and 10k testing images of $28 \times 28$ of 10 classes. The last 10k of training images are considered as the validation set. The network designed for MNIST is shown in Figure \ref{fig:mnistNetwork}. 

Relu is used as nonlinearity in the fully connected layers, and at the end of the network, a Softmax layer is applied to compute probability of predicted classes.  As loss function, cross-entropy function is used. To train the network, we use batch size of 512 on a single GPU, and Adam optimizer \cite{kingma2014adam} is deployed with $\beta_1= 0.9$ and $\beta_2= 0.999$. Learning rate is initialized to $10^{-3}$ and divided by 10 after $100th$ and $300th$ epochs, we also reduce the learning rate gradually over epochs by 0.9995. The learning rate is reduced over batches in an epoch as well by 0.9995 for the first 100 epoch and by 0.99995 for the rest of training. The best model is selected based on the validation error. Images are normalized to be in the range of $[0,1]$; for training images, we apply a simple data augmentation by randomly shifting images horizontally and vertically ($10\%$). Table \ref{tab:comarResultsMNIST} shows the model performance for MNIST dataset.

\begin{table}
\parbox{.3\linewidth}{
\centering
\caption{Comparison results for MNIST dataset}
\label{tab:comarResultsMNIST}
\begin{small}
\begin{tabular}{lcc}
MNIST & Accuracy\\
\toprule
Our method &99.58\% & \\
state-of-the-art \cite{wan2013regularization}& 99.79\%\\
\cite{xu2015multi} &99.58\\
\cite{goodfellow2013maxout}&99.45\\
\bottomrule
\end{tabular}
\end{small}
}
\hfill
\parbox{.3\linewidth}{
\centering
\caption{Comparison results for CIFAR-10 dataset}
\label{tab:comarResultsCifar10}
\begin{small}
\begin{tabular}{lcc}
CIFAR-10 & Accuracy\\
\toprule
Our method &88.18\% & \\
state-of-the-art \cite{huang2018gpipe}& 99.0\%\\
\cite{visin2015renet} &87.65\%\\
\cite{krizhevsky2012imagenet}&89\%\\
\bottomrule
\end{tabular}
\end{small}
}
\hfill
\parbox{.3\linewidth}{
\centering
\caption{Comparison results for CIFAR-100 dataset}
\label{tab:comarResultsCifar100}
\begin{small}
\begin{tabular}{lcc}
CIFAR-100 & Accuracy\\
\toprule
Our method &63.31\% & \\
state-of-the-art \cite{huang2018gpipe}& 91.3\%\\
\cite{srivastava2013discriminative} &63.15\%\\
\cite{lin2013network}&64.39\%\\
\bottomrule
\end{tabular}
\end{small}
}
\end{table}

CIFAR-10 dataset  \cite{krizhevsky2009learning}consists 50k training images and 10k testing images of size $32 \times 32$ in 10 classes; the last 10k of the training images are considered as validation. The network designed for this dataset has 11 layers including FIO (\#rules=64, \#outputs=32, fuzzy sets= $3 \times 3$, stride=1) $\rightarrow$ FIO(\#rules=64, \#outputs=32, fuzzy sets=$3 \times 3$, stride=1)  $\rightarrow$ FPO(\#rules=128, \#outputs=64, fuzzy sets=$2 \times 2$, stride=2) $\rightarrow$ FIO(\#rules=128, \#outputs=64, fuzzy sets=$3 \times 3$, stride=1) $\rightarrow$ FIO(\#rules=128, \#outputs=64, fuzzy sets= $3 \times 3$, stride=1) $\rightarrow$  FPO(\#rules=128, \#outputs=64, fuzzy sets= $2 \times 2$, stride=2) $\rightarrow$ FIO(\#rules=256, \#outputs=128, fuzzy sets= $3 \times 3$, stride=1) $\rightarrow$ FPO(\#rules=256, \#outputs=128, fuzzy sets= $2 \times 2$, stride=2) $\rightarrow$ FL(\#units=512, dropout=0.2) $\rightarrow$ FL(\#units=512, dropout=0.2) $\rightarrow$ FL(\#units=10) where FIO, FPO and FL stand for fuzzy inference operation, fuzzy pooling layer and fully connected layer, respectively.

The model trained for MNIST is used as a pretraining model for this dataset. Leaky-Relu is used as nonlinearity for fully connected layers. For training the model, same batch size and optimizer as the MNIST model is used. The learning rate is initialized to $10^{-3}$ and divided by 10 when validation error plateaus. Between epochs, the learning rate is reduced by 0.9995 and between batches by 0.99994. The best set of parameters for testing is selected by validation error. We apply sample-wise subtraction and sample-wise standard deviation normalization on the training and testing images. For training images, we do a simple data augmentation by randomly flipping images horizontally and randomly shifting images horizontally and vertically ($20\%$). Table \ref{tab:comarResultsCifar10} shows the results.

CIFAR-100 has the same format as CIFAR-10 with 100 classes. Same as CIFAR-10, we use the last 10k images as validation set. The network designed for this dataset has 11 layers including FIO (\#rules=64, \#outputs=32, fuzzy sets= $3 \times 3$, stride=1) $\rightarrow$ FIO(\#rules=64, \#outputs=32, fuzzy sets=$3 \times 3$, stride=1) $\rightarrow$ FPO(\#rules=128, \#outputs=32, fuzzy sets= $2 \times 2$, stride=2) $\rightarrow$ FIO(\#rules=128, \#outputs=32, fuzzy sets= $3 \times 3$, stride=1) $\rightarrow$ FIO(\#rules=128, \#outputs=32, fuzzy sets=$3 \times 3$, stride=1) $\rightarrow$ FPO(\#rules=128, \#outputs=64, fuzzy sets=$2 \times 2$, stride=2) $\rightarrow$ FIO(\#rules=128, \#outputs=64, fuzzy sets=$3 \times 3$, stride=1)  $\rightarrow$ FPO(\#rules=128, \#outputs=64, fuzzy sets=$2 \times 2$, stride=2) $\rightarrow$ FL(\#units=512, dropout=0.2) $\rightarrow$ FL(\#units=512, dropout=0.2) $\rightarrow$ FL(\#units=10). We train the network same as CIFAR-10. Table \ref{tab:comarResultsCifar100} shows the comparison results.

Table \ref{tab:comarResultsMNIST}-\ref{tab:comarResultsCifar100} compare performance of our proposed method with the current state-of-the-arts on each of the benchmarks. We have also compared our model with the closest models in terms of accuracy. Table \ref{tab:comarResultsMNIST} shows our model performance for MNISTcompared to Maxout networks \cite{goodfellow2013maxout} and models based on multi-loss regularization \cite{xu2015multi}. Table \ref{tab:comarResultsCifar10} shows the model performance compared to recurren neural network \cite{visin2015renet} and AlexNet \cite{krizhevsky2012imagenet}. Table \ref{tab:comarResultsCifar100} shows our performance compared to models with tree-based priors \cite{srivastava2013discriminative} and network-in-network model \cite{lin2013network}.

Our preliminary experiments confirm that the deep neuro-fuzzy system performance is comparable to other deep network structures, and it can be a viable structure for image analysis. However, the model does not outperform state-of-the-art on any of the three datasets, which calls for further research on this topic in the future.

\section{Conclusions}

We have proposed the first end-to-end deep neuro-fuzzy network for image classification. The network is not pushing state-of-the-art results but shows that deep structures based on fuzzy models can be applicable in image analysis. In this paper, we only investigated design of two new operations based on TSK model. However, in future, development of more operations based on TSK model and other fuzzy models such as Mamdani model \cite{pedrycz2007fuzzy} can be studied for image classification. Moreover, more research on developing regularization methods based on fuzzy models can improve performance of future models.

\small

\bibliography{deep-fuzzy-sonbol_2019-05-21_final}
\bibliographystyle{ieeetr}

\end{document}